# TinyTracker: Ultra-Fast and Ultra-Low-Power Edge Vision In-Sensor for Gaze Estimation

Pietro Bonazzi[1]*, Thomas Rüegg[1]*, Sizhen Bian[3], Yawei Li[2], Michele Magno[3]

Center for Project-based Learning D-ITET, ETH Zürich, Switzerland

[1]{pbonazzi, rueeggth}@ethz.ch, [2] yawei.li@vision.ee.ethz.ch [3]{sizhen.bian, michele.magno}@pbl.ee.ethz.ch

*Abstract*—Intelligent edge vision tasks encounter the critical challenge of ensuring power and latency efficiency due to the typically heavy computational load they impose on edge platforms. This work leverages one of the first "Artificial Intelligence (AI) in sensor" vision platforms, IMX500 by Sony, to achieve ultra-fast and ultra-low-power end-to-end edge vision applications. We evaluate the IMX500 and compare it to other edge platforms, such as the Google Coral Dev Micro and Sony Spresense, by exploring gaze estimation as a case study. We propose TinyTracker, a highly efficient, fully quantized model for 2D gaze estimation designed to maximize the performance of the edge vision systems considered in this study. TinyTracker achieves a 41x size reduction ($\sim$ 600Kb) compared to iTracker [1] without significant loss in gaze estimation accuracy (maximum of 0.16 cm when fully quantized). TinyTracker's deployment on the Sony IMX500 vision sensor results in end-to-end latency of around 19ms. The camera takes around 17.9ms to read, process and transmit the pixels to the accelerator. The inference time of the network is 0.86ms with an additional 0.24 ms for retrieving the results from the sensor. The overall energy consumption of the end-to-end system is 4.9 mJ, including 0.06 mJ for inference. The end-to-end study shows that IMX500 is 1.7x faster than Coral Micro (19ms vs 34.4ms) and 7x more power efficient (4.9mJ VS 34.2mJ).

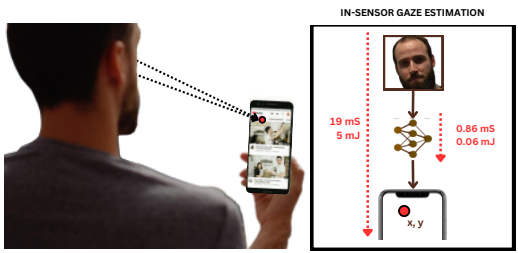

Fig. 1. **End-to-end gaze estimation on edge vision platforms with the state-of-the-art inference time (0.86 ms) and energy (0.06 mJ )**

FUNDING & ACKNOWLEDGMENTS

This research was funded by Innosuisse (103.364 IP-ICT). We thank A. Jaworowski and D.M. Arroyo for helpful conversations.

## I. INTRODUCTION

Deploying vision AI models at the battery-powered extreme edge poses significant challenges due to the computational heaviness of vision AI algorithms and the need for accurate and real-time inference [2]–[4]. This challenge becomes particularly critical considering the widespread application scenarios of edge vision AI in healthcare [5], personal assistants [6], wildlife monitoring [7], and more. In response, researchers have proposed compression techniques, such as quantization [8], [9], pruning [10], [11] and hardware design innovations [12]. However, achieving the optimal balance between accuracy and resource efficiency remains a key research focus in this field.

Emerging platforms like Sony IMX500 [13] embed AI in sensors, enabling real-time processing at the data source, reducing latency, enhancing privacy by eliminating data transmission. To leverage the power of this platforms, algorithm optimization, model compression, quantization are vital [14]–[19].

In this work, we propose a compressed sub-Mbyte vision model called Tinytracker, aiming to push the envelope of edge vision AI regarding power and latency efficiency. TinyTracker is designed for edge gaze estimation, which raises hard requirements of both energy efficiency and low latency, to work effectively in virtual reality, medical diagnosis, and assistive technologies [20].

The main contributions of this paper are: (1) proposing TinyTracker (Fig. 1) to demonstrate the feasibility of end-to-end milliseconds gaze estimation latency on novel in-sensors AI cores; (2) comparing and evaluating state-of-the-art commercial hardware solutions available for low-power, high-speed end-to-end computer vision tasks at the edge.

## II. RELATED WORK

### A. Tiny Machine Learning

Due to the growing interest in deep learning on edge devices, several recent studies have been conducted to benchmark deep learning algorithms on single-board computers and embedded platforms [21]–[23]. For example, Mozhgan et al. [24] explored a vision-based autonomous drone navigation system based on GAP8 MCU and achieved 40.6 ms latency with 34 mJ energy per inference, which shows the state-of-the-art performance compared to the existing edge vision solution on tiny drones [25]. However, to the best of our knowledge, a comprehensive comparative analysis of end-to-end (from sensing to processing) edge AI vision platforms is missing. We address this gap in this work by profiling three commercial vision hardware solutions in an end-to-end fashion, from the moment the image is captured to the neural network prediction: the Sony IMX500 [13] with a stacked camera and processor as one chip solution, Sony Spresense [26] with cable-connected Sony IMX500 image sensor, and Coral Dev Micro [27] with Himax HM01B0 CMOS sensor.

*These authors contributed equally to this work.

*B. Gaze Estimation Algorithms*

Gaze estimation [1], [28]–[32], has traditionally relied on facial geometry. However, these methods struggle under varying lighting, head pose, and rapid eye movements. Event cameras [33] promise high-frequency gaze estimation, but their high cost and limited availability restrict the wide deployment. Recently, Convolutional Neural Networks (CNNs) [1], [34], [35], Vision Transformers (ViT) [31], and Capsule Networks [32] have successfully learned the image-gaze mapping, by training on large-scale datasets (i.e., GazeCapture [1], ETH-XGaze [36], etc.), these networks have shown good generalization capabilities and increased precision.

However, these deep learning solutions have not yet been optimized for edge devices with resources and real-time performance constraints. We adress this gap and introduce TinyTracker, an efficient network capable of accurate gaze prediction from images with only 2.5 cm error, maintaining low power (0.06 mJ/Inference), and an inference latency of 0.86 ms, making it ideal for resource-constrained scenarios.

## III. TINYTRACKER: IN-SENSOR GAZE ESTIMATION

Our proposed model, TinyTracker, based on iTracker [1], is designed to operate within the constraints of edge devices. The original model iTracker is a CNN that requires four inputs: face and eye images and a face grid, all extracted by a face detection algorithm. However, for edge devices, this multiple-input design is not feasible, due to its complexity, lack of support, and higher memory requirements. Consequently, we streamline TinyTracker by eliminating eye and face grid inputs. To compensate for the lost face location data, we concatenate the face image coordinates to the input (grid embedding) and use greyscale images to maintain three-channels on the input. In essence, TinyTracker incorporates a MobileNetV3 backbone [37] pre-trained on ImageNet [38], with an added convolutional layer and two fully connected layers. It reduces parameter count by 13.8x and Multiply-Accumulate (MAC) operations by 224.7x compared to the baseline, as documented in Tab. I.

TABLE I
MODEL COMPARISON

| Name | Input res. | Params | MAC | Size[MB] |
|---|---|---|---|---|
| iTracker (Baseline) | 224x224 | 6'287k | 2651M | 24.6 |
| TinyTracker (Ours) | 112x112 | 455k | 11.8 M | 0.6 |

We train the model for up to 20 epochs on GazeCapture [1], using an NVIDIA RTX 3070. We use the Adam optimizer at a learning rate of 0.001 and a batch size of 64. The training data is augmented by adding noise and applying random contrast, saturation, and hue adjustments. After training the network is quantized to 8-bit integers while retaining 32 floating-point precision on the outputs.

## IV. HARDWARE PLATFORMS

As mentioned, this work leverages the novel IMX500 to perform in-sensor gaze estimation. Moreover, we profile two other edge platforms: the Coral Dev Micro [27], and Sony Spresense [26] as seen in Fig. 2.

*1) Sony Spresense:* The Spresense main board features an ARM Cortex-M4F CPU with 6 cores, running at a maximum frequency of 156 MHz. It offers 1.5 MB of Static Random-Access Memory (SRAM) and 8 MB of flash memory. The board includes a dedicated parallel interface for camera input [39].

*2) Coral Dev Micro:* The Coral Dev Micro features an ARM Cortex-M7, ARM Cortex-M4 and a Coral Edge TPU ML accelerator which provides 4 TOPS at 2 watts of power with 128 MiB NAND flash and 64 MB of Synchronous Dynamic Random-Access Memory (SDRAM) and a maximum clock frequency of 500Mhz. The board contains a built-in color camera and a PDM microphone.

*3) Sony IMX500:* The Sony IMX500 is an advanced image sensor designed for edge AI applications. It features a stacked pixel architecture and a built-in AI processor, which eliminates the need for external memory or high-performance processors. The pixel chip captures information across a wide angle with 12.3 effective megapixels, while the logic chip performs high-speed AI processing.

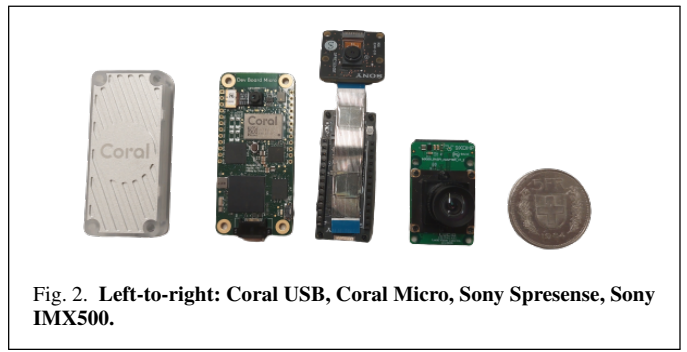

Fig. 2. **Left-to-right: Coral USB, Coral Micro, Sony Spresense, Sony IMX500.**

## V. EVALUATION METRICS

*A. Hardware performance*

To supply a fair benchmarking, we evaluate our model on images of size 112x112 pixels on the following metrics:
1) **Total Latency:** [ms] End-to-end time measuring image capture and inference.
2) **Total Energy:** [mJ] Energy consumed by hardware during image capture and inference.
3) **Inference Efficiency:** [MAC/Cycle] Measures parallelism of a given hardware.
4) **Latency:** [ms] Time required to calculate inference.
5) **Energy (E) per Inference:** [mJ] Energy consumed by hardware during a single inference

6) **Power Efficiency (P):** [mW/MHz] Power consumption normalized in terms of clock frequency.

The resolution is chosen to fit within the memory limitations of all hardware platforms.

### B. Model performance

The model performance is evaluated for both float32 and int8 models, in terms of the gaze prediction error in centimeters, and follows the same evaluation procedure as in [1].

## VI. RESULTS

### A. Hardware Performance

The evaluation results for Spresense [26], IMX500 [13] and Coral Micro [27] can be found in Fig. 3 and Tab. II. Additionally, we provide measurements on the Coral USB accelerator.

TABLE II
HARDWARE EVALUATION

| Platform | Spresense | CoralUSB | CoralMicro | IMX500 |
|---|---|---|---|---|
| End-to-End Evaluation ||||| 
| E [mJ]↓ | 234.1 | - | 34.2 | **4.9** |
| Latency [ms] ↓ | 522.5 | - | 34.4 | **19** |
| Inference Evaluation ||||| 
| MAC/Cycle ↑ | 0.20 | 54.09 | 8.69 | **73.23** |
| Latency [ms] ↓ | 386.60 | 0.87 | 5.43 | **0.86** |
| P [μW/MHz]↓ | 530.13 | 4436.40 | 5553.20 | **274.58** |
| E [mJ] ↓ | 31.97 | 0.97 | 6.02 | **0.06** |

Comparing inference efficiency, both IMX500 and Coral display parallel processing capabilities, with IMX500 being more efficient than Coral (73.23 vs 8.69 MAC/Cycle). Spresense trails significantly, achieving only 0.20 MAC/Cycle due to its reliance on a single core for inference. Coral (USB) and IMX500 perform almost identically in terms of inference speed (0.87 ms vs 0.86 ms). However, on the Coral Micro DevBoard, the inference time increases to 5.43 ms due to additional I/O processing. In the end-to-end evaluation, IMX500 dramatically reduces processing time, yielding an end-to-end latency of a mere 19.0 ms in comparison to Spresense (522.5 ms) and Coral (34.4 ms). This can be attributed to its unique design, which facilitates direct loading of images into the AI accelerator hardware.

Concerning energy consumption per inference, IMX500 significantly outperforms, requiring only 0.06 mJ. In contrast, Spresense and Coral consume 31.97 mJ and 0.97 mJ (USB)/6.02 mJ (Micro) respectively. Despite Coral's higher power consumption, its efficiency per inference improves by shorter inference latency than Spresense. The distinction in energy consumption becomes more pronounced when considering the end-to-end. Spresense's high energy consumption (234.1 mJ) results from the camera remaining active post-image capture. Although Coral deactivates the camera, its edge TPU module still consumes a substantial amount of energy during idle periods resulting in a total of 34.2 mJ. IMX500, on the other hand, only activates its dedicated hardware as needed, thus conserving energy during idle phases resulting in an energy consumption of 4.9 mJ.

Power efficiency, expressed in $\mu$W/MHz, reveals IMX500 leading with a value of 274.58. Spresense and Coral follow with 530.13 $\mu$W/MHz and 4436.4 (USB)/ 5553.2 (Micro) $\mu$W/MHz respectively.

### B. Model Performance

Table III compares TinyTracker to its predecessor, iTracker, using the checkpoint from the official GitHub repository [40]. We trained the model on 1.2M images and cross-validated on around 200k (15%) samples. The error rates are similar, with TinyTracker slightly outperforming it by 0.12 cm on the original resolution and under-performing by 0.08 cm on reduced resolution. Quantization affects the model similarly at both resolutions. Our model comparison for both RGB and Greyscale with grid embedding (G) inputs reveals that adding localization information improves prediction precision, in our case by 0.5 cm, aligning with [1]'s findings, which uses a face grid for the same purpose.

TABLE III
PRECISION COMPARISON *MODEL USING BATCHNORM

| Res | Model | Error [cm] | Error int8 [cm] |
|---|---|---|---|
| 224 | iTracker | 2.46 | - |
| | TinyTracker (G) | 2.34 | 2.63 |
| 112 | iTracker | 2.40* | - |
| | TinyTracker (G) | 2.54 | 2.62 |
| | TinyTracker RGB | 2.90 | 3.07 |

## VII. CONCLUSION

This paper evaluated the first 'AI in sensor' platform. We introduced TinyTracker, an efficient model for 2D gaze estimation on edge vision systems. TinyTracker achieves a remarkable 41x size reduction (600KB) while maintaining high precision. Even when fully quantized, the loss in precision is only 0.16 cm. On the Sony IMX500 vision sensor, TinyTracker has an end-to-end latency of 19 ms and consumes 4.9 mJ. The Sony IMX500 outperforms Sony Spresense by 27.5x in speed and is 20x more power-efficient than the Coral Edge TPU.

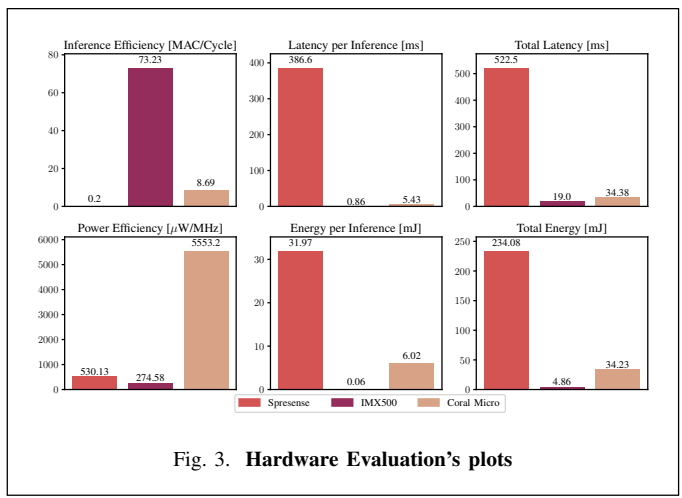

Fig. 3. **Hardware Evaluation's plots**

Our findings emphasize the importance of sensor-integrated AI accelerators and the effectiveness of tiny machine learning for scalable computer vision designs.